\def\year{2019}\relax
\documentclass[letterpaper]{article} 
\usepackage{aaai19}  
\usepackage{times}  
\usepackage{helvet}  
\usepackage{courier}  
\usepackage{url}  
\usepackage{graphicx}  
\graphicspath{ {Images/}}
\usepackage{amsfonts}
\usepackage{algorithm, algorithmicx, algpseudocode, float}
\usepackage{amsmath, amssymb}
\usepackage{mathtools}
\usepackage{ntheorem}
\usepackage{color}
\usepackage{array}
\usepackage{multirow}
\usepackage{float}
\usepackage{booktabs}
\usepackage{comment}
\usepackage{booktabs,subcaption,amsfonts,dcolumn}
\usepackage{tabu}
\usepackage{siunitx}
\usepackage[normalem]{ulem}
\newtheorem{lemma}{{\bf Lemma}}

\begin{document}
%
\title{On the Persistence of Clustering Solutions and\\ True Number of Clusters in a Dataset}
\author{Amber Srivastava$^{1a}$, Mayank Baranwal$^{2b}$, Srinivasa Salapaka$^{1c}$\\
$^1$University of Illinois at Urbana-Champaign, $^2$University of Michigan, Ann Arbor\\
$^a$asrvstv6@illinois.edu, $^b$mayankb@umich.edu, $^c$salapaka@illinois.edu}
\maketitle
\begin{abstract}
Typically clustering algorithms  provide clustering solutions with prespecified number of clusters. The lack of a priori knowledge on the true number of underlying clusters in the dataset makes it important to have a metric to compare the  clustering solutions with different number of clusters. This article  quantifies a notion of  {\em persistence} of clustering solutions that enables comparing solutions with different number of clusters.  The persistence relates to the range of data-resolution scales over which a clustering solution persists; it is quantified in terms of the maximum over two-norms  of all the associated  cluster-covariance  matrices. Thus we associate a persistence value for each element in a set of clustering solutions with different number of clusters. We show that the datasets where {\em natural} clusters are a priori known, the clustering solutions that identify the natural clusters are most persistent - in this way, this notion can be used to identify solutions with {\em true} number of clusters. Detailed experiments on a variety of standard and synthetic datasets demonstrate that the proposed persistence-based indicator outperforms the existing approaches, such as, gap-statistic method, $X$-means, $G$-means, $PG$-means, dip-means algorithms and information-theoretic method, in accurately identifying the clustering solutions with true number of clusters. Interestingly, our method can be explained in terms of the phase-transition phenomenon in the deterministic annealing algorithm, where the number of distinct cluster centers changes (bifurcates) with respect to an annealing parameter.
\end{abstract}

\section{Introduction and Related Work}\label{sec:Intro}
Many high-impact application areas such as bio-informatics \cite{andreopoulos2009roadmap}, \cite{di2008evolutionary}, exploratory data mining \cite{larose2014discovering}, combinatorial drug discovery \cite{sharma2008scalable}, data and network aggregation \cite{yuan2014data}, medical imaging \cite{huang2015intuitionistic} and many other information processing fields have fueled significant work on clustering algorithms.  Most of these algorithms such as $k$-means \cite{hartigan1979algorithm}, $k$-medoids \cite{park2009simple}, and expectation-maximization \cite{dempster1977maximum} require the number of clusters to be prespecified. Despite substantial work on clustering algorithms, there is relatively scant literature on determining the true number of clusters in a dataset. In this context, it should be noted that there is no single agreed-upon notion of \emph{natural} clusters or {\em true} number of clusters; typically existing algorithms make assumptions on the datasets (e.g. generated from a mixture of Gaussian distributions) and validate their results on datasets that satisfy the assumptions.

There are various measures developed to characterize the clustering solutions resulting from a clustering algorithm with different number ($k$) of clusters. One of the popular methods for determining the number of clusters is based on computing \emph{gap statistic}~\cite{tibshirani2001estimating}. It compares the total intracluster variation for different values of $k$ (number of clusters) with their expected values under null reference distribution of the data. The number of clusters $k$ is ascribed to the case where the gap is largest. However, as remarked in \cite{feng2007pg}, this method works well for finding a small number of clusters, but has difficulty as the true $k$ increases.

Some of the recent methods that determine the number of clusters under some assumptions on datasets include - $\quad \quad X$-means algorithm \cite{pelleg2000x}, where clustering using $k$-means is performed for a range of number $k$ of clusters, and the value $k_t:=k$ that yields the  best Bayesian Information Criterion (BIC) \cite{kass1995reference} score is chosen as an estimate for the true number of clusters. Other related algorithms use criteria such as Akaike information criteria \cite{akaike2011akaike} or minimum description length \cite{rissanen1978modeling} instead of BIC. The $X$-means algorithm works well for well-separated spherical clusters but tends to overfit in the case of non-spherical clusters \cite{feng2007pg}. 

The information-theoretic approach \cite{sugar2003finding} where it estimates the number of true clusters $k_t$ by detecting a significant {\em jump} in the modified distortion $D^{\gamma}$ vs $k$ plot; here $D$ is the clustering distortion objective, $k$ is the number of clusters, $\gamma \approx -d/2$, and the data points are in $\mathbb{R}^d$.  Although the choice $\gamma \approx -d/2$ works well for certain datasets, one can   find examples where this choice fails \cite{sugar2003finding}. 
\begin{figure*}[t]
  \begin{center}
    \begin{tabular}{cc}
    	\includegraphics[width=1\columnwidth]{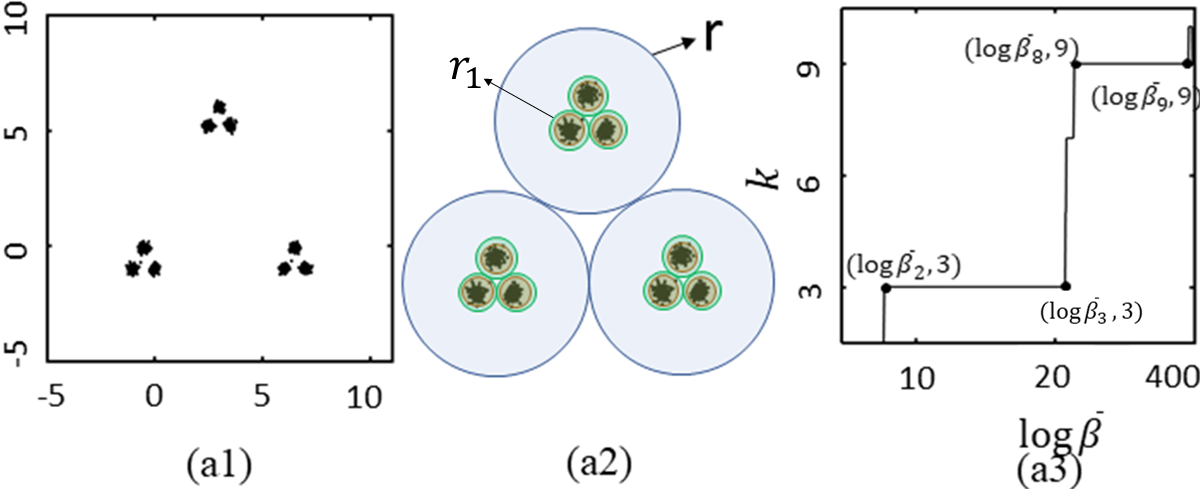} &
     \includegraphics[width=1\columnwidth]{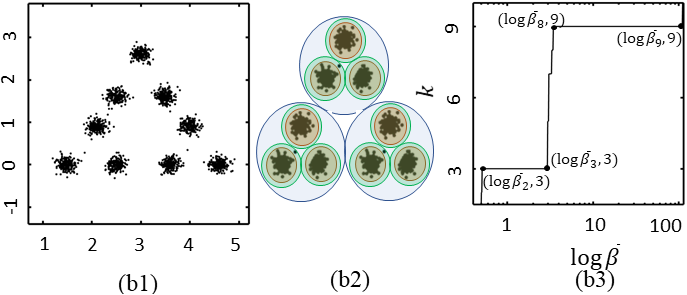}\\
    \end{tabular}
  	\caption{ Illustration of a mixture of nine Gaussian distributions arranged in groups of three superclusters. In (a1) the three superclusters are well separated from each other while in (b1) they are closer to each other. Observe that in (a2) for a large range in resolution scales (radii $r$) within the blue annulus, each supercluster appears as a single cluster, and only for a small range of resolution scale depicted by green annulus, each Gaussian distribution is identifiable separately. In other words for a large range of resolution the only the three superclusters are distinguishable from one another while for a smaller range of resolution each Gaussian distribution is identified separately. In (b1) since the three superclusters are closer to each other, the range of resolution scales within the blue annulus gets reduced, thereby indicating existence of three natural clusters in (a1) and nine natural clusters in (b1).}\label{fig: resolution}
  \end{center}
\end{figure*}
$G$-means \cite{hamerly2004learning} algorithm identifies the number of clusters in a dataset under the assumption that each cluster is a Gaussian distribution. It is a hierarchical algorithm that increases the number of clusters $k$ until the hypothesis that each cluster comes from a single Gaussian distribution is validated; typically done using the Anderson-Darling statistic test \cite{stephens1974edf} on each cluster after projecting it onto a one-dimensional space. $PG$-means algorithm \cite{feng2007pg} is an improvement on the $G$-means algorithm, where the number of clusters in a Gaussian mixture model is obtained by applying the Kolmogorov-Smirnov (KS) test to the one-dimensional projection of the \emph{entire} dataset;  $PG$-means also works well when the true clusters are overlapping with each other. 

Dip-means \cite{kalogeratos2012dip} is another method that assumes the dataset is generated from a mixture of unimodal distributions. Here, Hartigan's dip statistic test \cite{hartigan1985dip} is used to verify the unimodal nature of the admissible cluster. The authors also extend the dip-means algorithm to shape clustering problems by using kernel $k$-means \cite{dhillon2004kernel} for clustering. Other alternative approaches in the literature to estimate the true number of clusters are Bayesian $k$-means \cite{kurihara2009bayesian} that uses Maximization-Expectation to learn a mixture model, a method based on repairing faults in Gaussian mixture models \cite{sand2001repairing} and various stability-based model validation methods \cite{lange2003stability}, \cite{tibshirani2005cluster}, \cite{levine2001resampling}. The main drawbacks in most of the above existing methods stem from the underlying restrictive assumptions on the datasets; accordingly, the algorithms do not perform well when datasets do not meet the assumptions, which is often the case when considering standard non-synthetic datasets. These methods fail to accurately estimate the true number of clusters in most of the standard datasets as illustrated in the experiment section in this paper. 

In this article we develop a notion of {\em persistence} of clustering solutions that enables comparing solutions, which result from a clustering algorithm,  with different number of clusters. Here we do not make any assumptions on the underlying data distribution. Since a clustering solution requires grouping a set of points in such a way that points in the same cluster are more similar to each other than to those in other clusters. We characterize {\em persistence} of a clustering solution as the {\em range} of resolution scales for which (a) points within each cluster seem indistinguishable, and (b) points in different clusters are distinguishable. For instance, Figure \ref{fig: resolution}(a1) illustrates a dataset containing nine Gaussian clusters which are arranged in groups of three superclusters. If we choose the resolution scale of radius $r$, as shown in Figure \ref{fig: resolution}(a2), then the points within each super-cluster is indistinguishable. Therefore, one will conclude that at this resolution level the dataset consists of only three clusters. Also note that in Figure \ref{fig: resolution}(a1), the three super-clusters are persistent for a large range of resolution scales as indicated by the thickness of the blue annulus around each of them. On the other hand, the green annulus around each of the nine Gaussian clusters, is relatively thinner indicating that a clustering solution that identifies all the nine Gaussian clusters is relatively less persistent. 

In a later section we quantify this notion of persistence of a clustering solution with $k$ distinct clusters. In particular, the persistence is characterized in terms of the maximum over two-norms of all the cluster covariance matrices at two successive values of $k$. We also show analytically how for a clustering solution that identifies natural clusters, this measure correctly estimates the true number of clusters through a simple illustrative example consisting of spherical clusters with uniform distributions. We also provide extensive experimental results on a variety of standard and synthetic datasets in a later section. The results demonstrate that our method outperforms over the existing algorithms described above. In particular, our method correctly estimates the true number of clusters on 13 of the 14 benchmark datasets tested, whereas the next best method could estimate the true number of clusters only on 7 benchmark instances.

\section{Persistence of a Clustering Solution and its Quantification}\label{sec: PersistClusters}

{\bf Intuitive description : } The notion of a cluster can be related to the resolution scales at which a dataset is viewed. For instance, on one hand the entire dataset can be considered as a single cluster, while on the other hand each data point can also be considered as a cluster by itself. Thus in Figure \ref{fig: resolution}(a1), at low resolution scale (characterized here  by a radius greater than the diameter of the entire dataset) no two points of the dataset are  distinguishable from each other and the entire dataset is deemed a single cluster. Now consider a clustering solution at a higher resolution scale (for instance resolution characterized by the radius $r$ in Figure \ref{fig: resolution}(a2)), the datapoints from the three superclusters become distinct from one another and we are able to identify these as three distinct clusters in the dataset. Upon further increasing the resolution scale (for instance resolution characterized by the radius $r_1$) in Figure \ref{fig: resolution}(a2), the datapoints sampled from each of the nine Gaussian distributions become distinct from one another and we are able to identify the nine clusters in the dataset. On further increasing the resolution scale, each point by itself will be regarded as a cluster.

We use resolution to capture the notion of persistence of clustering solution. We propose that the clustering solution that persists for large range of resolution is a good indicator of natural clusters and corresponding number estimates true number of clusters. In fact, after quantifying persistence later in this section, we show that for both the datasets in Figure \ref{fig: resolution}(a1) and \ref{fig: resolution}(b1), the persistence is larger for clustering solutions with three and nine clusters, while they are relatively small for clustering solutions with other number of clusters.  Moreover, we observe that the three super-clusters are more persistent than the nine clusters for the dataset in Figure \ref{fig: resolution}(a1), while nine clusters are more persistent than the three super-clusters in Figure \ref{fig: resolution}(b1); this is also intuitive from the relative thickness of the blue and green annuli in Figures \ref{fig: resolution}(a2) and \ref{fig: resolution}(b2). Therefore, this suggests that the three super-clusters are \emph{more} natural in Figure \ref{fig: resolution}(a1) while the nine clusters are {\em more} natural in Figure \ref{fig: resolution}(b1). These inferences agree with our intuition from visual inspection of these datasets and are also corroborated by the measure proposed later in this section.

Let $\beta_k$ denote the lowest resolution scale at which $k+1$ clusters are identifiable in a dataset. We define the persistence of a clustering solution with $k$ clusters as $[\log \beta_{k}-\log \beta_{k-1}]$. The {\em true number} of clusters can be estimated in terms of persistence of clustering solution. Accordingly, if the clustering into $k$ groups is persistent for a long range of resolution scales, without $k+1$ clusters becoming evident in the dataset, then $k$ is a good estimate for the true number of clusters. In other words, for a clustering solution with true number $k_t$ of natural clusters it takes a large change in the resolution scale for the data points, originally belonging to the same cluster, to become distinguishable enough so as to belong to different clusters. Our hypothesis is that for a clustering solution with $k_t$ natural clusters $\log \beta_{k_t} - \log \beta_{k_t-1} > \log \beta_{k} - \log \beta_{k-1}$ for all $k \neq k_t$.

{\bf Quantification of persistence of a clustering solution and resolution scales:} This quantification is substantially motivated by our reinterpretation of the deterministic annealing (DA) algorithm \cite{rose1998deterministic}. The DA algorithm mimics the annealing procedure studied in statistical physics literature and gives high quality clustering solutions on linearly separable data. We do not describe the DA algorithm in detail here albeit re-interpret the auxiliary cost function (referred to as {\em free-energy} in \cite{rose1998deterministic}) in DA to discern a possible use of its annealing parameter as a measure of resolution.

The DA algorithm views the problem of clustering a dataset $\mathcal{X} = \{x_i:x_i\in \mathbb{R}^d, 1\leq i\leq N\}$ consisting of $N$ data points into $k$ groups of nearly similar entities as an equivalent facility location problem (FLP), where the goal is to allocate a set of facilities $\mathcal{Y}=\{\mathbf{y_j}:y_j\in\mathbb{R}^d, 1\leq j\leq k\}$ to data points $\{\mathbf{x_i}\}$ such that the cumulative distance between data points and their nearest facilities is minimum. Note that the facility locations are indeed the centroids of individual clusters (for linearly separable data and with squared Euclidean metric) and the FLP viewpoint is critical to many commonly used clustering algorithms, such as $k$-means . Thus, a solution to FLP results in clustering of the underlying dataset where the corresponding clusters $\{\pi_j\}$ are defined by {\em voronoi partitions} 
$\pi_j=\{\mathbf{x_i}\in \mathcal{X} : \mathbf{y_j}=\text{arg}\min_{\{\mathbf{y_l}\}}d(\mathbf{x_i},\mathbf{y_l})\}$. More precisely, we consider the following optimization problem for FLP
\begin{align}\label{eq:FLP}
D = \min\limits_{\mathcal{Y}} \sum\limits_{i=1}^Np_i\sum\limits_{\{\mathbf{y_j}\}}\min\limits_{1\leq j\leq k}d(\mathbf{x_i},\mathbf{y_j}),
\end{align}
where $p_i$ denotes a known relative weight of vector $\mathbf{x_i}$ (e.g. $p_i=\frac{1}{N}$), and $d(\mathbf{x_i},\mathbf{y_j})$ is a measure of distance between $\mathbf{x_i}$ and $\mathbf{y_j}$ which is usually considered to be the squared Euclidean distance. In data compression literature $D$ is usually referred to as the distortion function \cite{gersho2012vector}. DA considers the log-sum-exp approximation where it approximates $ \min\limits_{1\leq j\leq k}d(\mathbf{x_i},\mathbf{y_j})$ by $-\frac{1}{\beta}\log \sum_{j=1}^k e^{-\beta d(\mathbf{x_i},\mathbf{y_j})}$, which results in the following smooth optimization problem that approximates (\ref{eq:FLP})
\begin{align}\label{eq: smooth_App}
F = \min\limits_{\mathcal{Y}}-\frac{1}{\beta}\sum_{i=1}^N p_i \log \sum_{\{\mathbf{y_j}\}} e^{-\beta d(\mathbf{x_i},\mathbf{y_j})},
\end{align}
which is parameterized by $\beta \in \mathbb{R}$. The parameter $\beta$ determines the extent of approximation of $D$ by $F$. At larger values of $\beta \rightarrow \infty$ the approximation function $F$ tends to converge at the distortion $D$. On the other hand, at low values of $\beta$ $(\approx 0)$ approximation $F$ is considerably distinct from the distortion $D$. Here $F$ is referred to as the {\em free-energy} function and $\beta$ as an annealing parameter in the DA algorithm. We obtain the minimum (local) of $F$ at a given $\beta$, by setting the partial derivative $\frac{\partial F}{\partial \mathbf{y_j}}$ to zero, which results in the following centroid-like condition for squared Euclidean distances:
\begin{align}\label{eq:Y}
	\mathbf{y_j} = \big(\sum\limits_{i=1}^Np_ip(j|i)\mathbf{x_i}\big)\Big/\big(\sum\limits_{i=1}^Np_ip(j|i)\big), \text{where} \\ p(j|i) = \big(e^{-\beta d(\mathbf{x_i},\mathbf{y_j})}\big)\Big/\big(\sum\limits_{\mathbf{y_j}\in\mathcal{Y}}e^{-\beta d(\mathbf{x_i},\mathbf{y_j})}\big).
\end{align} 

We justify the annealing parameter $\beta$ as a measure of resolution as follows. Note that for any two data points $\mathbf{x_1}$ and $\mathbf{x_2}$ in the bounded dataset,  the term $e^{-\beta d(\mathbf{x_1},\mathbf{y_j})} \approx e^{-\beta d(\mathbf{x_2},\mathbf{y_j})}$ when $\beta$ is small ($\beta\approx 0$), i.e. points $\mathbf{x_1}$ and $\mathbf{x_2}$ are indistinguishable. More precisely, for every $\epsilon>0$, there exists $\beta>0$ small enough such that $\left|e^{-\beta d(\mathbf{x_1},\mathbf{y_j})}- e^{-\beta d(\mathbf{x_2},\mathbf{y_j})}\right|<\epsilon$. Note that from (\ref{eq:Y}), at small values of $\beta$ $(\approx 0)$ all the facilities $\{\mathbf{y_j}\}$ are coincident. We can deduce from here that at low $\beta$ values, no two data points are distinguishable (within $\epsilon$), that is, the optimization problem (\ref{eq: smooth_App}) cannot differentiate between them, and therefore entire dataset will be deemed as a single cluster. In fact, the DA algorithm associates only one resource $\mathbf{y_1}$ at the centroid of the entire dataset. Now as $\beta$ increases, two distinct points that originally belonged to the same cluster $\pi_j$, become distinguishable for large enough $\beta$, i.e. $e^{-\beta d(\mathbf{x_1},\mathbf{y_j})}$ no longer approximates $e^{-\beta d(\mathbf{x_2},\mathbf{y_j})}$; thus the problem (\ref{eq: smooth_App}) can differentiate between these data points, and they need not necessarily belong to the same cluster. In the limit $\beta \rightarrow \infty$, all the data points are entirely distinct from each other and the optimal solution to (\ref{eq: smooth_App}) is to assign a distinct facility to each point since no two data points are similar enough to be put in the same cluster. Thus $\beta$ quantifies our intuitive notion of resolution. Equivalently, $\log \beta$ quantifies the resolution scale.  

To quantify {\em persistence} of a clustering solution in the FLP setup, we need to determine the range of resolution scales $\log \beta$ over which a clustering solution persists. Note that a clustering solution $\{\pi_j\}$ to the FLP persists till the set of cluster centers $\mathcal{Y}=\{\mathbf{y_j}\}$ cease to be a minima of $F$ as the annealing parameter $\beta$ (resolution) is increased. Therefore in context of the relaxed problem (\ref{eq: smooth_App}), we need to compute $\mathcal{Y}$ such that it minimizes $F$ and find the range of values of $\log \beta$ for which it remains a minimum, i.e. at every resolution level in this range the optimal centers $\mathcal{Y}$ must satisfy
\begin{align}\label{eq:saddle1}
	\frac{d}{d\epsilon}F(\mathcal{Y}+\epsilon\Psi)\big|_{\epsilon=0} = 0, \ \ \text{and} 
\end{align}
\begin{align}\label{eq:saddle2}
 H(\mathcal{Y},\Psi,\beta):=\frac{d^2}{d\epsilon^2}F(\mathcal{Y}+\epsilon\Psi)\big|_{\epsilon=0} > 0,
\end{align}
for all finite perturbations $\Psi$.
The cluster centers $\mathcal{Y}$ ceases to be a minimum of $F$ for a value of $\beta$ when the Hessian (\ref{eq:saddle2}) is no longer positive definite, that is when there exists a perturbation $\Psi$ such that $H(\mathcal{Y},\Psi,\beta)$ is no longer positive definite. Now it can be shown that for $d(x_i,y_j)$ as squared Euclidean distance the Hessian is
\begin{small}
\begin{align}\label{eq:bifurcation}
H(\mathcal{Y},\Psi,\beta)&=\sum\limits_{\mathbf{y_j}}p_ip(j|i)\boldsymbol{\psi_j}^T\left[I-2\beta  \ C_{\mathcal{X}|\mathbf{y_j}}^{k}\right]\boldsymbol{\psi_j}\nonumber\\
&+ \sum\limits_{i=1}^Np_i\Big[\sum\limits_{\mathbf{y_j}}p(j|i)(\mathbf{x_i}-\mathbf{y_j})^T\boldsymbol{\psi_j}\Big]^2 ,
\end{align}
\end{small}
\begin{align}\label{eq: CCM}
\text{where }
C_{\mathcal{X}|\mathbf{y_j}}^{k}\triangleq\sum\limits_{i=1}^Np(i|j)(\mathbf{x_i}-\mathbf{y_j})(\mathbf{x_i}-\mathbf{y_j})^T
\end{align}
is the \emph{cluster covariance matrix} of the posterior distribution $p(i|j)$ and $I$ is the identity matrix of appropriate dimensions. From (\ref{eq:bifurcation}), it is not difficult to show that $H(\mathcal{Y},\Psi,\beta)$ loses its positivity only when $\text{det}\left[I-2\beta\ C_{\mathcal{X}|\mathbf{y_0}}^k\right] = 0$ at some $\mathbf{y_0}\in \mathcal{Y}$ (please refer to supplementary material for proof)\footnotemark\footnotetext{arXiv:1811.00102}; therefore the  critical value of $\beta$ beyond which the clustering center $\mathcal{Y}=\{\mathbf{y_j}\in\mathbb{R}^d:1\leq j \leq k\}$ is no longer a minimum is given by
\begin{align}\label{eq: critical_beta}
	\beta_k = {1}\Big/\big(2\lambda_\text{max}(C_{\mathcal{X}|\mathbf{y_0}}^k)\big),
\end{align}
where $\lambda_{\max}(C_{\mathcal{X}|\mathbf{y_0}}^k)$ is the largest eigenvalue of $C_{\mathcal{X}|\mathbf{y_0}}^{k}$. In fact $\mathbf{y_0}$ is the centroid of that cluster which has the maximum variance, i.e. $\mathbf{y_0}=\arg\max_{\{\mathbf{y_j}\}} \lambda_{\max}(C_{\mathcal{X}|\mathbf{y_j}}^{k})$. Therefore beyond the critical value $\beta_k$ in (\ref{eq: critical_beta}) the number of identifiable cluster increases by one so as to identify a new minimum of $F$ \cite{rose1998deterministic}. This makes intuitive sense, since we would expect the clusters with biggest variance to split before the others. The spread directions are indicated by the associated eigenvectors of the covariance matrix, with the largest spread along the eigenvector corresponding to the largest eigenvalue. Therefore using this analysis we can quantify persistence of a clustering solution with $k$ clusters by $v(k):=\log \beta_{k}-\log\beta_{k-1}$, where $\beta_k$, is the resolution at which the number of distinct clusters increases from $k-1$ to $k$. The  {\em true} number $k_t$ of clusters can be estimated by $\text{arg}\max_k v(k)$. 
\begin{figure*}[t]
  \begin{center}
  	\includegraphics[width=0.65\textwidth]{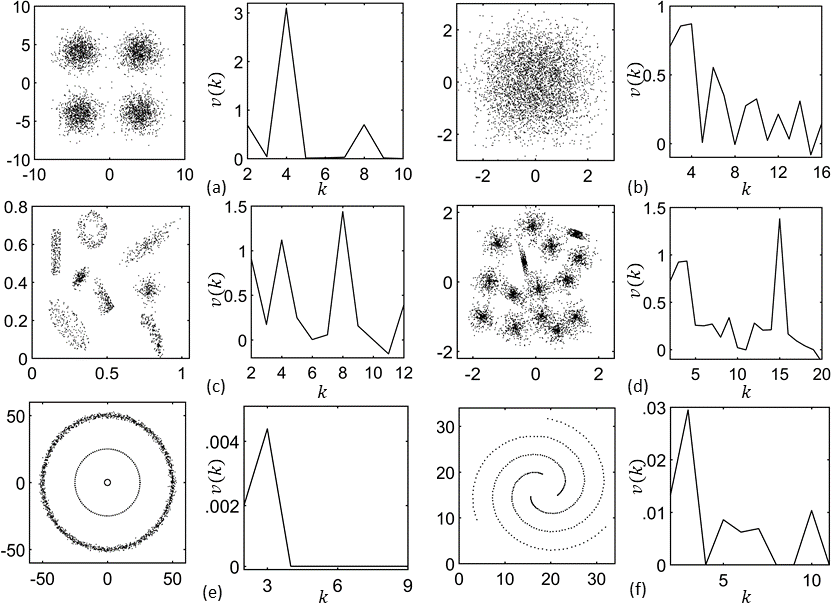}
  \end{center}
  \caption{Evaluation of our method on a variety of synthetic datasets - (a) Low variance Gaussian, $k_t = 4$, (b) High variance Gaussian, $k_t = 4$, (c) ComboSetting, $k_t = 8$, (d) Synthetic-15 (S15), $k_t = 15$, (e) Concentric rings, $k_t = 3$, and (f) Spirals, $k_t = 3$. Our method predicts the correct number of clusters in each of these scenarios.}
  \label{fig:eval}
\end{figure*}

{\bf Making the persistence independent of the DA algorithm:} Note that, though our quantification of persistence of a clustering solution is motivated from the DA algorithm, we can easily make it algorithm independent by replacing  soft associations in (\ref{eq: CCM}) with hard associations; thus we re-define the cluster covariance matrix (\ref{eq: CCM}) for clustering solution with $k$ clusters as
\begin{align}\label{eq: CCM_redefine}
\bar{C}_{\mathcal{X}|\mathbf{y_j}}^k = \sum_{i=1}^N \nu_{ij}(\mathbf{x_i}-\mathbf{y_j})(\mathbf{x_i}-\mathbf{y_j})^T
\end{align}
where the posterior distribution term $p(i|j)\in[0,1]$ is replaced by $\nu_{ij}\in\{0,1\}$ that represents hard-associations between the vector $\mathbf{x}_i$ and cluster centroid $\mathbf{y}_j$ such that $\nu_{ij}=1$ if vector $\mathbf{x}_i$ belongs to the cluster $\pi_j$ and zero otherwise. We formally define the true number of clusters $k_t$ in a dataset as
\begin{align}\label{eq: K_def}
k_t:= \arg\max_k \Big[v(k):=\log \bar{\beta}_{k} - \log \bar{\beta}_{k-1}\Big],
\end{align}
\begin{align}\label{eq: crit_beta}
\text{where }
\bar{\beta}_{k} := \Big[\max\limits_{1\leq j\leq k} \big[2\lambda_{\max}\big(\bar{C}_{\mathcal{X}|\mathbf{y_j}}^k\big) \big]\Big]^{-1}
\end{align}
is the minimum resolution level at which $k+1$ distinct clusters centroids should be allocated to the dataset. 

\textbf{Remark on Scalability and use of $\log$ in $v(k)$: } Note that our proposed method is scalable with respect to size ($N$) of the dataset as it requires computation of the largest eigenvalue of a $d\times d$ covariance matrix $\bar{C}_{\mathcal{X}|\mathbf{y_j}}^k$ which can be computed in $O(d^2)$, where $d$ is the dimension of the feature-space. Our notion of persistence captures, by what factor one should scale resolution to obtain more clusters. This factor is simply expressed as a difference using log for better visualization. Also the range of resolutions is typically very large (each data point is a highest resolution cluster to the entire dataset being the lowest resolution cluster); therefore log function discriminates this range better.

\begin{algorithm}\label{alg: alg1}
	\caption{main($\mathcal{X}$,$k_{\max}$)}
    \begin{enumerate}
    \item[1.] Initialize $k=1$.
    \item[2.] Run a clustering algorithm on $\mathcal{X}$ with $k$ clusters and compute $\bar{\beta}_{k}$ using (\ref{eq: crit_beta}). 
    \item[3.] $k\longleftarrow k+1$. Go to step 2. Stop if $k = k_{\max}+1$.
    \item[4.] Choose $k_t$ using (\ref{eq: K_def}).
    \end{enumerate}
\end{algorithm}

Figures \ref{fig: resolution}(a3) and \ref{fig: resolution}(b3) illustrates our method for determining number of true clusters on two datasets considered in Figures \ref{fig: resolution}(a1) and \ref{fig: resolution}(b1) respectively. Observe that the quantities $\log \bar{\beta}_3-\log \bar{\beta}_2 \gg \log \bar{\beta}_{k}-\log \bar{\beta}_{k-1}$ and $ \log \bar{\beta}_{9} -\log\bar{\beta}_8 \gg \log \bar{\beta}_{k}-\log \bar{\beta}_{k-1}$ for all $k\neq 3$ and $k\neq 9$ in both the figures. Further, we observe in the Figure \ref{fig: resolution}(a3) that $\log \bar{\beta}_3 - \log \bar{\beta}_2 > \log\bar{\beta}_{9}-\log\bar{\beta}_8$ which indicates $k_t=3$ (using (\ref{eq: CCM_redefine})) for the dataset in Figure \ref{fig: resolution}(a1). Similarly we observe in Figure \ref{fig: resolution}(b3) that $\log\bar{\beta}_{9} - \log\bar{\beta}_8 > \log \bar{\beta}_3 - \log\bar{\beta}_2$ which indicates $k_t=9$ for the dataset in Figure \ref{fig: resolution}(b1). Again, these inferences agree with our intuition from visual inspection of these datasets. 
\begin{figure}
\centering
\includegraphics[width=0.50\columnwidth]{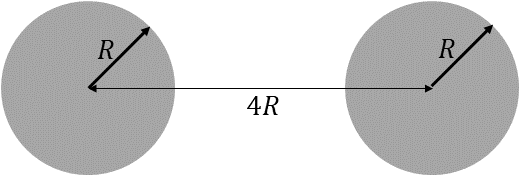}
\caption{Illustrates two circular clusters of radius $R$ with uniformly distributed data-points.}\label{fig: example}
\end{figure}
\begin{table*}[t]
	\sisetup{table-format=-1.2}
    \caption{Comparing algorithms on a variety of standard and synthetic datasets}\label{tab:results}
    \centering
    \begin{tabular}{lccccccc}
            \toprule
            \multirow{3}{*}{Algorithm} & {Low} & {High} & {Combo-} & &\\
            & {Variance} & {Variance} & {Setting} & {Wisconsin} &{Yeast} & {Glass} & {Leaves}\\
            \cmidrule(lr){2-8}
            &{$k_t=4$} &{$k_t = 4$} &{$k_t=8$} &{$k_t=2$} &{$k_t=10$} &{$k_t=6$} & {$k_t=100$}\\
            \midrule
            gap-statistic & \textbf{4} & 1 & 27 & 12 & 49 & 39 & 117\\
            $X$-means  & 1 &  1 & 25 & 6 & 47 & 1 & 219\\
            $G$-means  &  \textbf{4} &  68 & 33 & 83 & 56 & 15 & 66\\
            $PG$-means  & \textbf{4} & 3 & 12 & 6 & 3 & 7 & Error\\
            dip-means  & \textbf{4} &  1 & \textbf{8} & 11 & 1 & 1 & 3\\
            Info. Th.  & \textbf{4} & 3 & \textbf{8} & \textbf{2} & 2 & \textbf{6} & 99\\
            kernel dip-means  & - & - & - & - & - & - & -\\
            Our Method & \textbf{4} & \textbf{4} & \textbf{8} & \textbf{2} & \textbf{10} & \textbf{6} & \textbf{100}\\
            \toprule
            \multirow{3}{*}{Algorithm} & & & & & & {Concentric} & \\
            & {Wine} & {Iris} & {Banknote} & {Thyroid}& {Birch1} & {Rings} & {Spirals}\\
            \cmidrule(lr){2-8}
            &{$k_t=3$} &{$k_t = 3$} &{$k_t=2$} &{$k_t=3$}& {$k_t=100$} &{$k_t=3$} & {$k_t =3$}\\
            \midrule
            gap-statistic  & \textbf{3} & \textbf{3} & 58 & 17 & Error & n/a & n/a \\
            $X$-means  & 40 & 12 & 230 & 1 & 1 & n/a & n/a\\
            $G$-means  & 2 & 4 & 57 & 7 & 1953 & n/a & n/a\\
            $PG$-means  & 1 & 2 & 35 & \textbf{3} & 32 & n/a & n/a\\
            dip-means  & 1 & 2 & 4 & 1 & Error & n/a & n/a\\
            Info. Th.  & \textbf{3} & 2 & 5 & \textbf{3} & \textbf{100} & n/a & n/a\\
            kernel dip-means  & - & - & - & - & - &\textbf{3}  & 1\\
            Our method & \textbf{3} & 2 & \textbf{2} & \textbf{3} & \textbf{100} &\textbf{3} & \textbf{3}\\
            \bottomrule
		\end{tabular}
\end{table*}

{\bf Extending to nonlinearly separable data: }\label{sec:nonlin}
The method proposed above works well for linearly separable data. However, many problems such as shape clustering or clustering with pairwise distances often consist of data points that are not separable linearly. We use kernel trick \cite{dhillon2004kernel} to overcome this issue. Accordingly, we map data points $\mathbf{x_i}$ to an abstract higher-dimensional space (where data-points are linearly separable) through a suitably chosen kernel function $\phi(\cdot)$. While an explicit representation of kernel function $\phi(\cdot)$ is unknown, the inner-products $\phi(\mathbf{x_i})^T\phi(\mathbf{x_j})$ are known as elements of a kernel matrix $\mathcal{K}$ \cite{dhillon2004kernel}. Thus, in order to identify the clustering solution with the correct number of clusters in a nonlinearly separable dataset, one must evaluate the largest eigenvalue of the kernel data covariance matrix $\bar{C}_{\phi(\mathcal{X})|\mathbf{y_j}}^k$ defined as:
\begin{align}\label{eq:C_phi}
	\bar{C}_{\phi(\mathcal{X})|\mathbf{y_j}}^k = \sum\limits_{\mathbf{x_i}\in\pi_j}\left(\phi(\mathbf{x_i})-\mathbf{y_j}\right)\left(\phi(\mathbf{x_i})-\mathbf{y_j}\right)^T,
\end{align}
where $\pi_j$ denotes the $j$-th cluster. Computation of eigenvalues of $\bar{C}_{\phi(\mathcal{X})|\mathbf{y_j}}^k$ is not straightforward (since $\phi$ is unknown), however, we make use of the following lemma in order to obtain the spectral values of $\bar{C}_{\phi(\mathcal{X})|\mathbf{y_j}}^k$. 

\begin{lemma}\label{lemma:lemma1}
	Let $\bar{C}_{\phi(\mathcal{X})|\mathbf{y_j}}^k$ be the cluster covariance matrix as defined in (\ref{eq:C_phi}). Then $\bar{C}_{\phi(\mathcal{X})|\mathbf{y_j}}^k$ and $A=[A_{kl}]$ share the same non-zero eigenvalues, where $A_{kl} = \left(\phi(\mathbf{x_k})-\mathbf{y_j}\right)^T\left(\phi(\mathbf{x_l})-\mathbf{y_j}\right)$.  
\end{lemma}
\textbf{Proof}: Please refer to the supplementary material\footnotemark[\value{footnote}] for proof of the above lemma. 

Note that from (\ref{eq:Y}), the cluster center $\mathbf{y_j}=\dfrac{1}{|\pi_j|}\sum_{\mathbf{x_i}\in\pi_j}\phi(\mathbf{x_i})$. Thus elements of $A$ are known in terms of the elements of kernel matrix, and hence the spectral values of $\bar{C}_{\phi(\mathcal{X})|\mathbf{y_j}}^k$ can be easily obtained. 

We demonstrate the efficacy of our proposed metric using simulations on synthetic and standard datasets in the experiments section. Additionally, we analytically solve for the persistence of clustering solutions for an example problem as shown in Figure \ref{fig: example}. The Figure illustrates two equally sized circular clusters ($k_t=2$) with uniform distribution. One can easily compute the cluster covariance matrices for clustering solutions (as given by k-means) at various $k$'s and show that $v(2) > \eta$ $\forall$ $\eta \in \{v(3), v(4), v(5), v(6)\}$; thereby implying that for the dataset in Figure \ref{fig: example}, $k=2$ is a {\em more} natural choice of the number of clusters than $k=3,4,5 \text{ and } 6$. Please refer to the supplementary material\footnotemark[\value{footnote}] for the proof.
\begin{figure*}
\centering
\includegraphics[width=0.65\textwidth]{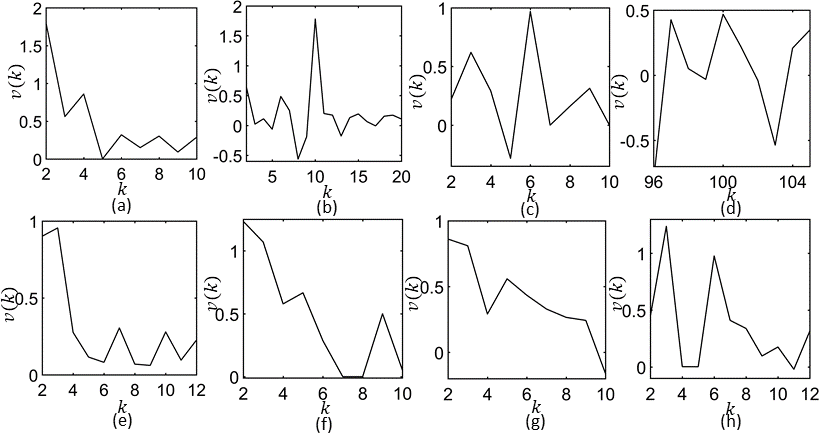}
\caption{Illustration of performance of our method on high-dimensional datasets - (a) Wisconsin ($d=9$, $N=681$, $k_t=2$), (b) Yeast ($d=8$, $N=1484$, $k_t=10$), (c) Glass ($d=9$, $N=214$, $k_t=6$) (d) Leaves ($d=64$, $N=1600$, $k_t=100$) (e) Wine ($d=13$, $N=178$, $k_t=3$) (f) Iris ($d=4$, $N=150$, $k_t=3$) (g) Banknote Authentication ($d=4$, $N=1372$, $k_t=2$) (h) Thyroid ($d=5$, $N=215$, $k_t=3$) . Note that, except for the Iris dataset, $v(k_t)$ is maximum in all the plots.}
    \label{fig:more-results}
\end{figure*}
\section{Experiments}\label{sec: Experiments}
In this section we employ the notion of persistence in estimating the true number of clusters in synthetic as well as standard datasets from the literature \cite{Dua:2017}, \cite{ClusteringDatasets} and provide comparisons with the gap-statistic method \cite{tibshirani2001estimating}, dip-means \cite{kalogeratos2012dip}, $X$-means \cite{pelleg2000x}, $G$-means \cite{hamerly2004learning}, $PG$-means \cite{feng2007pg} algorithms and the information theoretic approach \cite{sugar2003finding}. We observe that our proposed method outperforms the existing methods on various standard datasets as well as on synthetic datasets with acute overlap between two clusters.

As the method proposed in this paper evaluates a clustering solution for its persistence, any suitable clustering algorithm can be employed to obtain these clustering solutions. Note that irrespective of the criteria or metric used, a stable and good clustering solution at each $k$ is a prerequisite to correctly estimating the true number of clusters in a dataset. Since the $k$-means algorithm is ubiquitous in the data science literature, in all our simulations on linearly separable datasets we use $k$-means algorithm with multiple runs to determine the clustering solution at each value of $k$. For the purpose of estimating the true number of shape clusters we use the spectral clustering algorithm \cite{ng2002spectral} to obtain the clustering solutions at various $k$'s. Also note that for all the simulations demonstrated in this section we normalize every dataset to mean zero and standard deviation one.

Figure \ref{fig:eval} demonstrates multiple instances of synthetic datasets, and the corresponding plots of $v(k)$ versus $k$, where $k$ is the choice for the number of clusters. Figure \ref{fig:eval}a illustrates a mixture of four well separated Gaussian distributions. The corresponding plot of $v(k)$ versus $k$  shows a clear peak at $k=4$. Figure \ref{fig:eval}b illustrates another mixture of four Gaussian distributions with very high variances as compared to the Figure \ref{fig:eval}a. As can be seen, the sampled data is highly overlapping and it seems that the entire dataset is sampled from a single distribution. As seen in the corresponding $v(k)$ versus $k$ plot in the Figure \ref{fig:eval}b, our method exhibits a maximum value of $v(k)$ at $k=4$. On the other hand, the algorithms such as $G$-means, $PG$-means and dip-means fail to estimate the true number of clusters in this case (as shown in Table \ref{tab:results}), even though this dataset satisfies the assumptions required by these algorithms. In Figure \ref{fig:eval}c, the dataset is a mixture of well separated eight non-uniform clusters. The corresponding plot of $v(k)$ versus $k$ determines two distinguishable peaks at $k=4$ and $k=8$, although the peak is larger at $k=8$ and denotes the true number of clusters in this dataset.

Similar conclusions are observed for the linearly separable S15 dataset (a mixture of 15 Gaussian distribution \cite{Ssets}) in Figure \ref{fig:eval}(d), and other standard high-dimensional datasets - Wisconsin ($d=9$, $N=681$, $k_t=2$), Yeast ($d=8$, $N=1484$, $k_t=10$), Glass ($d=9$, $N=214$, $k_t=6$), Leaves ($d=64$, $N=1600$, $k_t=100$), Wine ($d=13$, $N=178$, $k_t=3$), Iris ($d=4$, $N=150$, $k_t=3$), Banknote Authentication ($d=4$, $N=1372$, $k_t=2$), Thyroid ($d=5$, $N=215$, $k_t=3$), shown in Figure \ref{fig:more-results},  where our proposed method estimates the true number of clusters appropriately. In particular, note that our proposed metric estimates the true number of clusters for a very high-dimensional leaves ($d=64$, $k_t=100$) dataset and a large Birch1 ($N=100,000$, $k_t=100$) dataset. All the other methods, except Information Theoretic approach on Birch1 dataset, fail to determine the true number of clusters in these two datasets.  Table \ref{tab:results} compares our method to the gap-statistic, $X$-means, $G$-means, $PG$-means, dip-means, kernel dip-means algorithms and information theoretic approach of determining the true number of clusters in the dataset. We observe that our method estimates the correct value of $k_t$ even for the high-variance Gaussian distribution Figure \ref{fig:eval}(b), yeast dataset, banknote authentication dataset and a high-dimensional Leaves dataset ($d=64$) where all the other methods fail to correctly estimate the true number of clusters. As previously noted, the proposed metric is scalable with respect to the size of the datasets and involves eigenvalue computation of a $d\times d$ cluster covariance matrix, where $d$ is the dimension of datapoints in the dataset. For the Iris dataset most of the methods, including ours, estimate the true number of clusters to be $2$. This is because the two of three clusters in the Iris dataset have significant overlap with each other. However, the gap-statistic method outperforms all the others and estimates correctly the true number of clusters in the Iris dataset.

As described earlier, our technique extends to non-linearly separable data too. Figure \ref{fig:eval}e and \ref{fig:eval}f illustrate two non-linearly separable datasets. We use the spectral clustering algorithm \cite{ng2002spectral} and determine the similarity graph using the Gaussian similarity function, $s(x_i,x_j)=\exp(-\|x_i-x_j\|^2/(2\sigma^2))$, where the $\sigma$ parameter is set to be $0.01$ and $0.08$ for (e) concentric rings, and (f) spirals, respectively. Table \ref{tab:results} shows comparison with the kernel dip-means \cite{kalogeratos2012dip} algorithm on these two non-linearly separable datasets. Note that the latter fails to correctly estimate the true number for clusters on the spiral dataset.

\section{Conclusion}
In this paper we study the persistence of a clustering solution both qualitatively and quantitatively. We use this persistence of clustering solutions to propose a simple yet effective approach for estimating the true number of clusters in a dataset. The key idea used here is to map the distinctiveness of number of clusters to phase-transition phenomenon occurring in the deterministic annealing algorithm. Moreover, we extend the results on linearly separable data to clustering of shapes (nonlinearly separable data) using kernel embedding. The proposed method does not make any assumptions on the underlying distributions that generate the dataset. Our simulations demonstrate the efficacy of this uncomplicated approach and experimental results shows that our method outperforms many other existing methods in literature.

\bibliography{ref}
\bibliographystyle{aaai}
\appendix
\section*{Supplementary material: On the number of clusters in data}

\section{Additional Experimental Results}\label{sec:additional}
\begin{figure}[H]
\begin{center}
\begin{tabular}{c}
\includegraphics[width=0.35\textwidth]{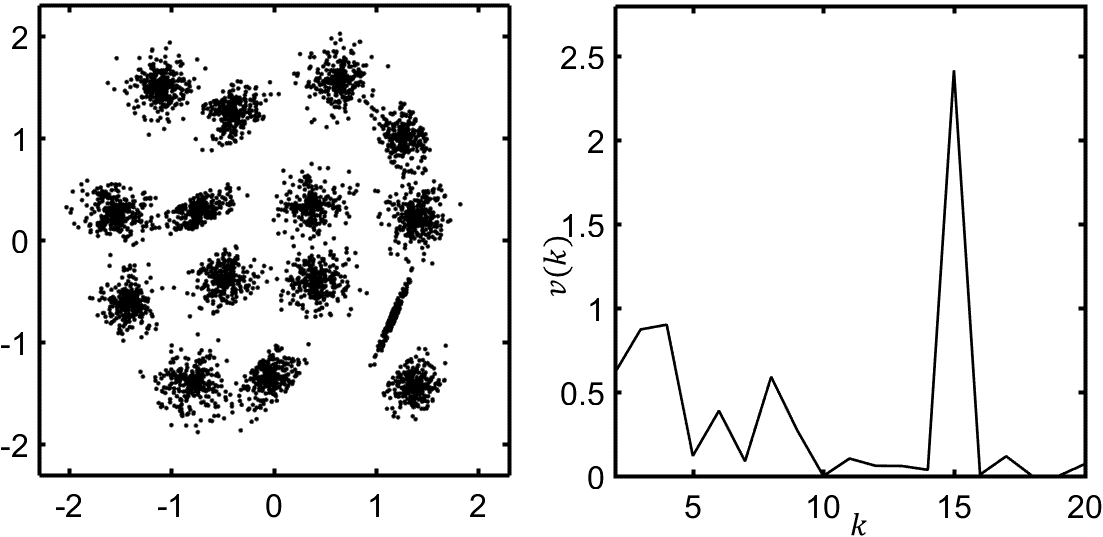}\\
(a) \\
\includegraphics[width=0.35\textwidth]{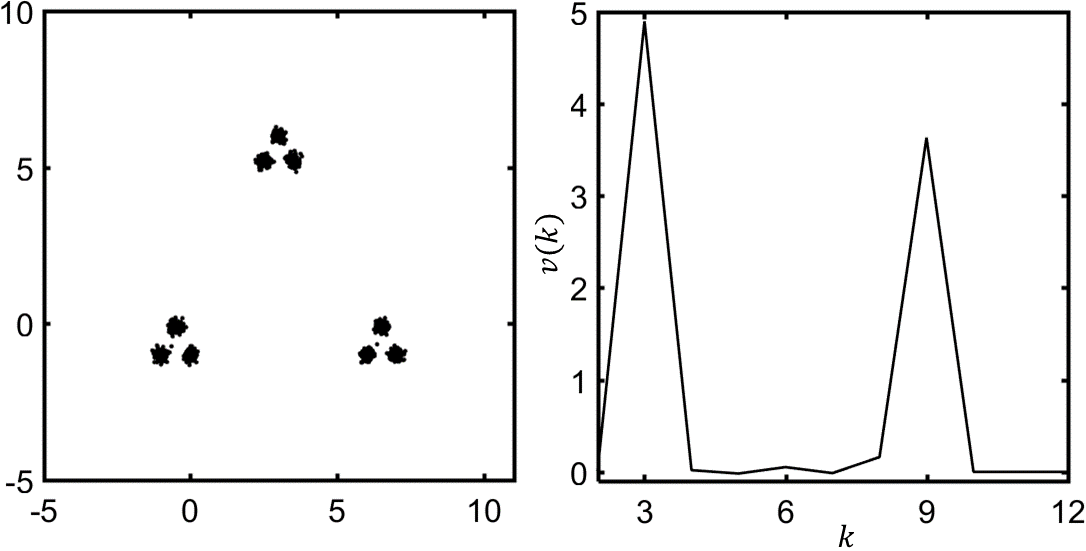}\\
(b)\\
\includegraphics[width=0.35\textwidth]{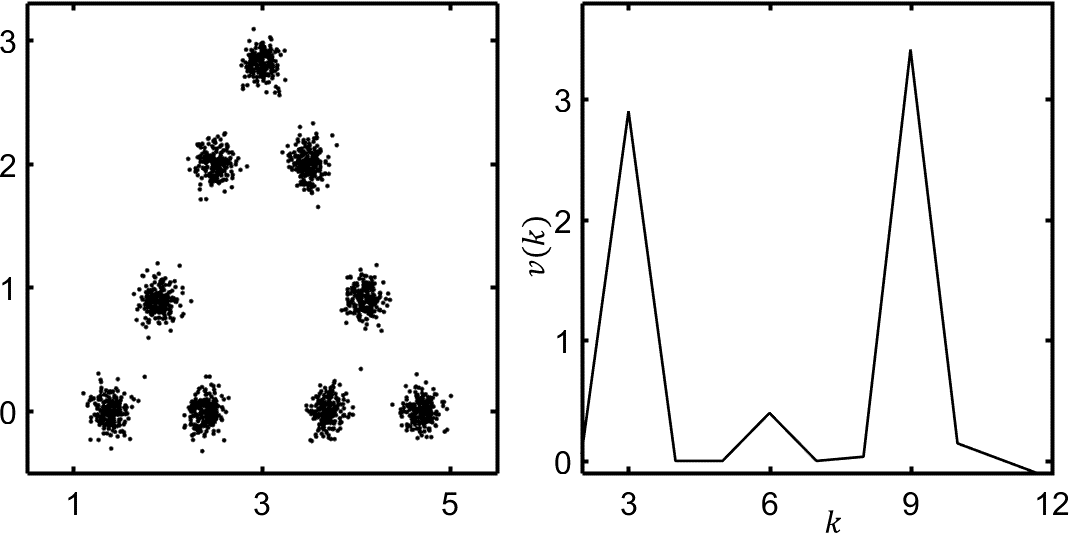}\\
(c) \\
\includegraphics[width=0.35\textwidth]{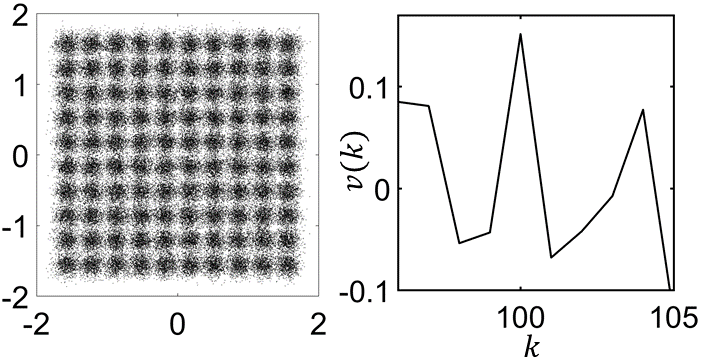}\\
(d)
\end{tabular}
\caption{(a) Synthetic 2-d data with $N=5000$ vectors and $k_t=15$ Gaussian clusters \cite{Ssets} with different degree of overlapping. The corresponding $v(k)$ versus $k$ plot for both the cases show a peak at $k=15$. (b) The $v(k)$ versus $k$ plot indicates $3$ to be true number of clusters. (c) The $v(k)$ versus $k$ plot indicates $9$ to be true number of clusters. (d) The $v(k)$ versus $k$ plot indicates $k_t=100$ for Birch1 dataset.}
\end{center}
\end{figure}
\section{Proof of Lemma1}\label{sec:proof}
{\bf Proof}: Let $(\lambda, \mathbf{u})$ be any eigenvector-eigenvalue pair of $A$. Here $\mathbf{u}\triangleq[u_1,u_2,\dots,u_N]$. Then, by definition:
\begin{align}\label{eq:lem1}
	&\sum\limits_{l=1}^NA_{kl}u_l = \lambda u_k \nonumber \\ 
    &\Rightarrow  \left(\phi(\mathbf{x_k})-\mathbf{y}\right)^T\sum\limits_{l=1}^N\left(\phi(\mathbf{x_l})-\mathbf{y}\right)u_l = \lambda u_k
\end{align}
Multiplying (\ref{eq:lem1}) by $\left(\phi(\mathbf{x_k})-\mathbf{y}\right)$ and then summing over $k$ yields:
\begin{align}\label{eq:lem2}
	\sum\limits_{k=1}^N&\left(\phi(\mathbf{x_k})-\mathbf{y}\right)\left(\phi(\mathbf{x_k})-\mathbf{y}\right)^T\underbrace{\sum\limits_{l=1}^N\left(\phi(\mathbf{x_l})-\mathbf{y}\right)u_l}_{\tilde{u}}\nonumber \\
    &= \lambda \underbrace{\sum\limits_{k=1}^N\left(\phi(\mathbf{x_k})-\mathbf{y}\right)u_k}_{\tilde{u}}, \nonumber \\
	&\Rightarrow C_{\phi(\mathcal{X})|\mathbf{y_j}}^k \ {\tilde{u}} = \lambda{\tilde{u}}
\end{align}
\begin{figure*}[t]
\centering
\includegraphics[width=\textwidth]{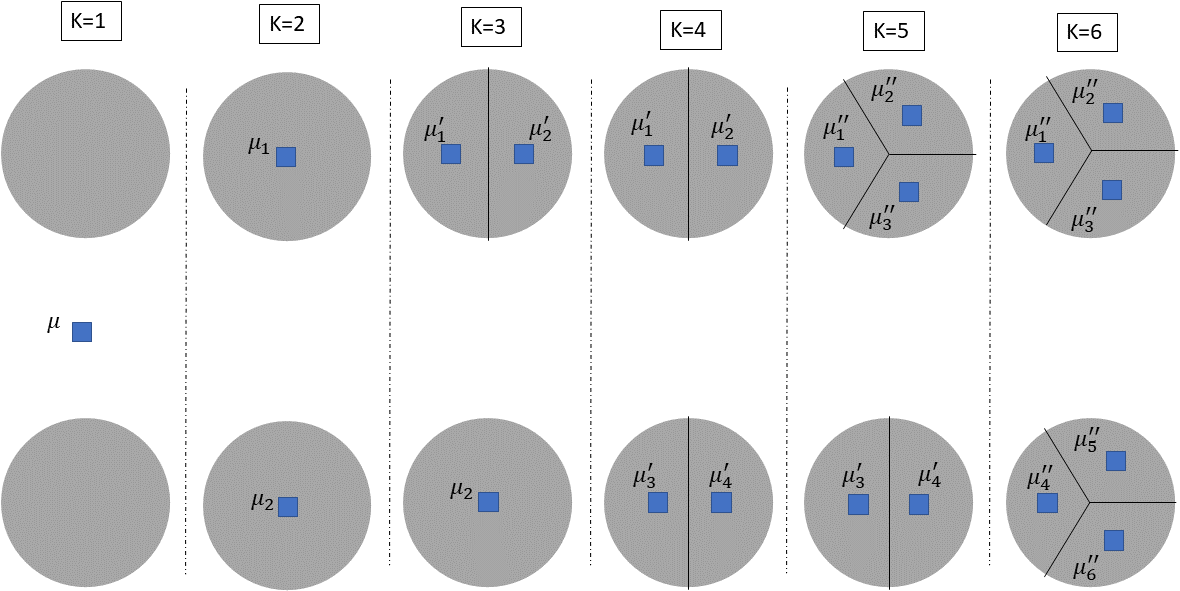}
\caption{The Figure illustrates the optimal clustering at each value of $k\in\{1,2,3,4,5,6\}$ for the dataset in the Figure \ref{fig: example}}\label{fig: ExampleProof}
\end{figure*}
\section{Proof for $H(\mathcal{Y},\Psi,\beta)=0$ only when det$(I-2\beta C_{\mathcal{X}|\mathbf{y_0}}^k)=0$}
We claim that $H(\mathcal{Y},\Psi,\beta)$ is non-negative for all finite perturbation $\Psi$ if and only if the matrix $[I-2\beta C_{\mathcal{X}|\mathbf{y_0}}^k]$ is positive definite. The 'if' part is straightforward since the second term in the expression is non-negative. For the 'only if' part we show that when $[I-2\beta C_{\mathcal{X}|\mathbf{y_0}}^k]$ is negative semi-definite definite, there exists a finite perturbation $\Psi$ such that the second term in $H(\mathcal{Y},\Psi,\beta)$ becomes zero thereby making the entire term negative. Let us assume that there exists a $\mathbf{y_0}\in\mathcal{Y}$ with positive probability such that the matrix $[I-2\beta C_{\mathcal{X}|\mathbf{y_0}}^k]$ is negative semi-definite. The perturbation $\Psi$ be such that $\Psi_{\mathbf{y}}=0$ $\forall$ $\mathbf{y}\neq\mathbf{y_0}$ and $\sum_{\mathbf{y}\in\mathcal{Y}:\mathbf{y}=\mathbf{y_0}}\Psi_{\mathbf{y}}=0$. Then the second term in $H(\mathcal{Y},\Psi,\beta)$ becomes zero. Thus whenever the first term in $H(\mathcal{Y},\Psi,\beta)$ is non-positive we can construct a perturbation such that the second term vanishes. Hence the positivity of the $H(\mathcal{Y},\Psi,\beta)$ for all perturbations $\Psi$ depends solely on the positive definiteness of $[I-2\beta C_{\mathcal{X}|\mathbf{y_0}}^k]$. The $\mathcal{Y}$ is no longer a minimum of $F$ when $H(\mathcal{Y},\Psi,\beta)=0$ which happens when the matrix $[I-2\beta C_{\mathcal{X}|\mathbf{y_0}}^k]$ loses its positive definiteness; i.e.
\begin{align*}
\det [I-2\beta C_{\mathcal{X}|\mathbf{y_0}}^k]=0
\end{align*}
The proof is as given in \cite{rose1998deterministic}.

\section{Proof for $v(2) > v(k)$ $\forall$ $k\in\{3,4,5,6\}$}
Figure \ref{fig: ExampleProof} illustrates an optimal clustering solution at various values of $k$. We assume that each circle contains $N$ datapoints. We have rotated the original dataset in Figure \ref{fig: example} by $90$ degrees for the ease of presentation; however, this will not change the corresponding problem and its solution. Note that Figure \ref{fig: ExampleProof} shows one of the optimal clusterings at each value of $k$; even if some other optimal clustering is chosen, due to symmetry, the cluster covariance matrix will remain unchanged. To evaluate $\bar{\beta}_k$ $\forall$ $k\in\{2,3,4,5,6\}$, we compute the relevant covariance matrices $\bar{C}_{\mathcal{X}|\mathbf{\mu}}^1$, $\bar{C}_{\mathcal{X}|\mathbf{\mu_1}}^2$, $\bar{C}_{\mathcal{X}|\mathbf{\mu_2}}^3$, $\bar{C}_{\mathcal{X}|\mathbf{\mu_1'}}^4$, $\bar{C}_{\mathcal{X}|\mathbf{\mu_3'}}^5$ and $\bar{C}_{\mathcal{X}|\mathbf{\mu_1''}}^6$.

\begin{figure}[ht]
\centering
\includegraphics[width=0.6\columnwidth]{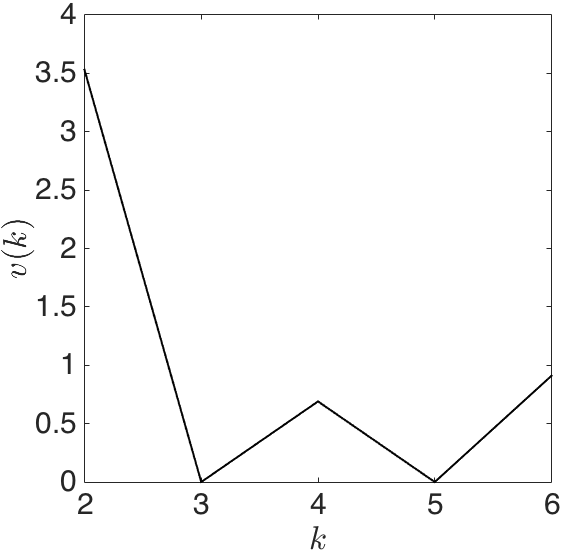}
\caption{Illustrates the $v(k)$ vs $k$ plot for the analytically calculated $v(k)$ at various $k$'s for the dataset in Figure \ref{fig: example}. As shown, $k=2$ is a more natural number of cluster than $k\in \{3,4,5,6\}$.}\label{fig: ExampleProofCluster}
\end{figure}

\begin{itemize}
\item Computing $\bar{C}_{\mathcal{X}|\mu_1}^2$.
\begin{align*}
\bar{C}_{\mathcal{X}|\mu_1}^2 &= \sum_{x\in C_{\mu_1}}(x-\mu_1)(x-\mu_1)^T \\
&\approx \int_{r,\theta}(x-\mu_1)(x-\mu_1)^T\rho r dr d\theta,\\
&\text{where}\quad x = [r\cos \theta \quad r\sin \theta]^T\\
&=\int_{0}^{2\pi}\int_{0}^R 
\begin{bmatrix}
r\cos \theta\\
r\sin \theta
\end{bmatrix}
\begin{bmatrix}
r \cos \theta & r\sin \theta
\end{bmatrix} \rho r dr d\theta\\
&\text{we assume $\mu_1=[0\quad 0]^T$ without loss of generality}\\
&=\int_{0}^{2\pi}\int_{0}^R
\begin{bmatrix}
r^2\cos^2\theta & r^2\cos\theta\sin\theta\\
r^2\sin\theta\cos\theta & r^2\sin^2\theta
\end{bmatrix}\rho rdrd\theta\\
&= \begin{bmatrix}
(\rho\pi R^2)R^2/4 & 0 \\
0 & (\rho\pi R^2)R^2/4
\end{bmatrix}\\
&\text{$(\rho\pi R^2)$ denotes the number of data points $N$}\\
&= \frac{NR^2}{4}\begin{bmatrix}
1 & 0 \\
0 & 1\\
\end{bmatrix}
\end{align*}
\begin{align*}
\Rightarrow \lambda_{\max}(\bar{C}_{\mathcal{X}|\mu_1}^2) = NR^2/4
\end{align*}
\item Computing $\bar{C}_{\mathcal{X}|\mu_2}^3$. Due to symmetry between the cluster $\pi_{\mu_1}$ and $\pi_{\mu_2}$ we can see that $\bar{C}_{\mathcal{X}|\mu_1}^2=\bar{C}_{\mathcal{X}|\mu_2}^3$. Hence
\begin{align*}
\Rightarrow \lambda_{\max}(\bar{C}_{\mathcal{X}|\mu_2}^3) = NR^2/4
\end{align*}
In Figure \ref{fig: ExampleProof}, we assume WLOG that $\pi_{\mu_1}$ splits and $\pi_{\mu_2}$ remains intact between $k=2$ and $k=3$.
\item Computing $\bar{C}_{\mathcal{X}|\mu}^1$
\begin{small}
\begin{align*}
\bar{C}_{\mathcal{X}|\mu}^1 &= \sum_{x\in C_{\mu}}(x-\mu)(x-\mu)^T\\
&=\sum_{x\in C_{\mu_1}}(x-\mu)(x-\mu)^T + \sum_{x\in C_{\mu_2}}(x-\mu)(x-\mu)^T\\
&=\sum_{x\in C_{\mu_1}}(x-\mu_1+(\mu_1-\mu))(x-\mu_1+(\mu_1-\mu))^T \\
&+\sum_{x\in C_{\mu_2}}(x-\mu_2+(\mu_2-\mu))(x-\mu_2+(\mu_2-\mu))^T\\
&= \bar{C}_{\mathcal{X}|\mu_1}^2 + \bar{C}_{\mathcal{X}|\mu_2}^2 + 2N\big(\frac{\mu_1-\mu_2}{2}\big)\big(\frac{\mu_1-\mu_2}{2}\big)^T\\
&\text{using the definition of $\bar{C}_{\mathcal{X}|\mu_1}^2$ and $\bar{C}_{\mathcal{X}|\mu_2}^2$}\\
&\text{and $\mu=\frac{\mu_1+\mu_2}{2}$}\\
&\text{note that } \mu_1-\mu_2 = [0\quad 4R]^T \\
&=\begin{bmatrix}
NR^2/2 & 0\\
0 & NR^2/2+ 8NR^2
\end{bmatrix}\\
&=NR^2\begin{bmatrix}
1 & 0 \\
0 & 8.5
\end{bmatrix}
\end{align*}
\end{small}
\begin{small}
\begin{align*}
\Rightarrow \lambda_{\max}({(\bar{C}_{\mathcal{X}|\mu}^1)})=8.5NR^2
\end{align*}
\end{small}
\item Computing $\bar{C}_{\mathcal{X}|\mu_1'}^4$. Since $\pi_{\mu_1'}$ is a semi-circle, we can easily locate the centroid $\mu_1'$ in it. Assuming that the origin $(0,0)$ is at the center of the full circle corresponding to this semi-circle, hence we have that $\mu_1'=[0 \quad \frac{4R}{3\pi}]^T$. 
\begin{small}
\begin{align*}
\bar{C}_{\mathcal{X}|\mu_1'}^4 &= \sum_{x\in C_{\mu_1'}} (x-\mu_1')(x-\mu_1')^T\\
&\approx \int_{0}^R\int_{0}^{\pi}\begin{bmatrix}
r\cos\theta\\
r\sin\theta-\frac{4R}{3\pi}
\end{bmatrix}
\begin{bmatrix}
r\cos\theta\\
r\sin\theta-\frac{4R}{3\pi}
\end{bmatrix}^T\rho rdrd\theta\\
&=NR^2\begin{bmatrix}
\frac{1}{8} & 0\\
0 & \frac{1}{8}+\frac{8}{9\pi^2}-\frac{16}{9\pi^2}
\end{bmatrix}
\end{align*}
\end{small}
\begin{small}
\begin{align*}
\Rightarrow \lambda_{\max}(\bar{C}_{\mathcal{X}|\mu_1'}^4)= 0.125NR^2
\end{align*}
\end{small}

\item Computing $\bar{C}_{\mathcal{X}|\mu_3'}^5$. Note that due to symmetry $\bar{C}_{\mathcal{X}|\mu_1'}^4=\bar{C}_{\mathcal{X}|\mu_3'}^5$.
\begin{align*}
\Rightarrow \lambda_{\max}(\bar{C}_{\mathcal{X}|\mu_3'}^5)=0.125NR^2
\end{align*}
\item Computing $\bar{C}_{\mathcal{X}|\mu_1''}^6$. Can be calculated as done above. We get
\[
\bar{C}_{\mathcal{X}|\mu_1''}^6=NR^2
\begin{bmatrix}
0.0165 & 0 \\
0 & 0.049
\end{bmatrix}
\]
\begin{align*}
\Rightarrow \lambda_{\max}(\bar{C}_{\mathcal{X}|\mu_1''}^6)=0.049NR^2
\end{align*}
\end{itemize}
Hence we have that $\bar{\beta}_1 = \frac{1}{17NR^2}$, $\bar{\beta}_2 = \frac{1}{2NR^2/4}$, $\bar{\beta}_3 = \frac{1}{2NR^2/4}$, $\bar{\beta}_4 = \frac{1}{0.125NR^2}$, $\bar{\beta}_5 = \frac{1}{0.125NR^2}$, $\bar{\beta}_6 = \frac{\pi}{0.98NR^2}$
\begin{align*}
v(2) &= \log \bar{\beta}_2-\log\bar{\beta}_1 = 3.53\\
v(3) &= \log \bar{\beta}_3-\log\bar{\beta}_2 = 0\\
v(4) &= \log \bar{\beta}_4-\log\bar{\beta}_3 = 0.69\\
v(5) &= \log \bar{\beta}_5-\log\bar{\beta}_4 = 0\\
v(6) &= \log \bar{\beta}_6-\log\bar{\beta}_5 = 0.91
\end{align*}
Hence we have that $v(2)>v(k)$ $\forall$ $k\in \{3,4,5,6\}$ as summarized in the Figure \ref{fig: ExampleProofCluster} 
\section{References to algorithmic implementation of various other metrics}
\begin{itemize}
\item Gap-statistic method: MATLAB in-built class {\em clustering.evaluation.GapEvaluation class}.
\item $X$-meas, $G$-means, dip-means, kernel dip-means: {\color{blue} http://kalogeratos.com/psite/material/dip-means/}
\item $PG$-means: Implementation provided by the author of this method \cite{feng2007pg}.
\item Information theoretic method: Self-implemented on MATLAB. 
\end{itemize}
\end{document}